\pdfoutput=1
\documentclass[letterpaper, 10 pt, conference]{ieeeconf}  

\IEEEoverridecommandlockouts                              
\usepackage{graphicx} 
\usepackage{multicol}
\usepackage{amsmath} 
\usepackage{amssymb}  
\usepackage{array}
\usepackage{hyperref}
\usepackage{balance}
\usepackage[detect-all]{siunitx}
\usepackage{booktabs}

\title{\LARGE \bf
The Open Vision Computer:\\An Integrated Sensing and Compute System for Mobile Robots
}

\author{Morgan Quigley$^{1}$, Kartik Mohta$^{2}$, Shreyas S. Shivakumar$^{2}$, Michael Watterson$^{2}$,
Yash Mulgaonkar$^{2}$, \\
Mikael Arguedas$^{1}$, Ke Sun$^{2}$,  Sikang Liu$^{2}$, Bernd Pfrommer$^{2}$, Vijay Kumar$^{2}$, Camillo J. Taylor$^{2}$
\thanks{We gratefully acknowledge the support of DARPA grants HR001151626 and HR0011516850.}
\thanks{$^{1}$Open Source Robotics Foundation. Mountain View, California \texttt{\{morgan, mikael\}@openrobotics.org}}
\thanks{$^{2}$GRASP Lab, University of Pennsylvania, Philadelphia PA 19104. \texttt{\{kmohta, sshreyas, wami, yashm, sunke, sikang, pfrommer, kumar, cjtaylor\}@seas.upenn.edu}}%
}

\begin{document}

\maketitle
\thispagestyle{empty}
\pagestyle{empty}

\begin{abstract}
In this paper we describe the Open Vision Computer (OVC) which was designed to support high speed, vision guided autonomous drone flight. In particular our aim was to develop a system that would be suitable for relatively small-scale flying platforms where size, weight, power consumption and computational performance were all important considerations. This manuscript describes the primary features of our OVC system and explains how they are used to support fully autonomous indoor and outdoor exploration and navigation operations on our Falcon 250 quadrotor platform.
\end{abstract}

\section{Introduction}

The primary goal of this project was to design an integrated sensing and computational system to support the demands of autonomous flight on smaller, lighter multi-rotor aircrafts, as shown in Figure~\ref{fig:Falcon_evolution}. In this effort, we seek to develop systems capable of flying through environments where GPS is commonly unavailable, such as under dense tree canopies or indoors. The resulting vision-guided systems require considerable onboard processing power while still controlling weight and overall system complexity. This paper describes the computational platform we developed, which we have named the Open Vision Computer (OVC).

\subsection{Motivation}
\begin{figure}[t]
	\centering
	\includegraphics[width=\linewidth]{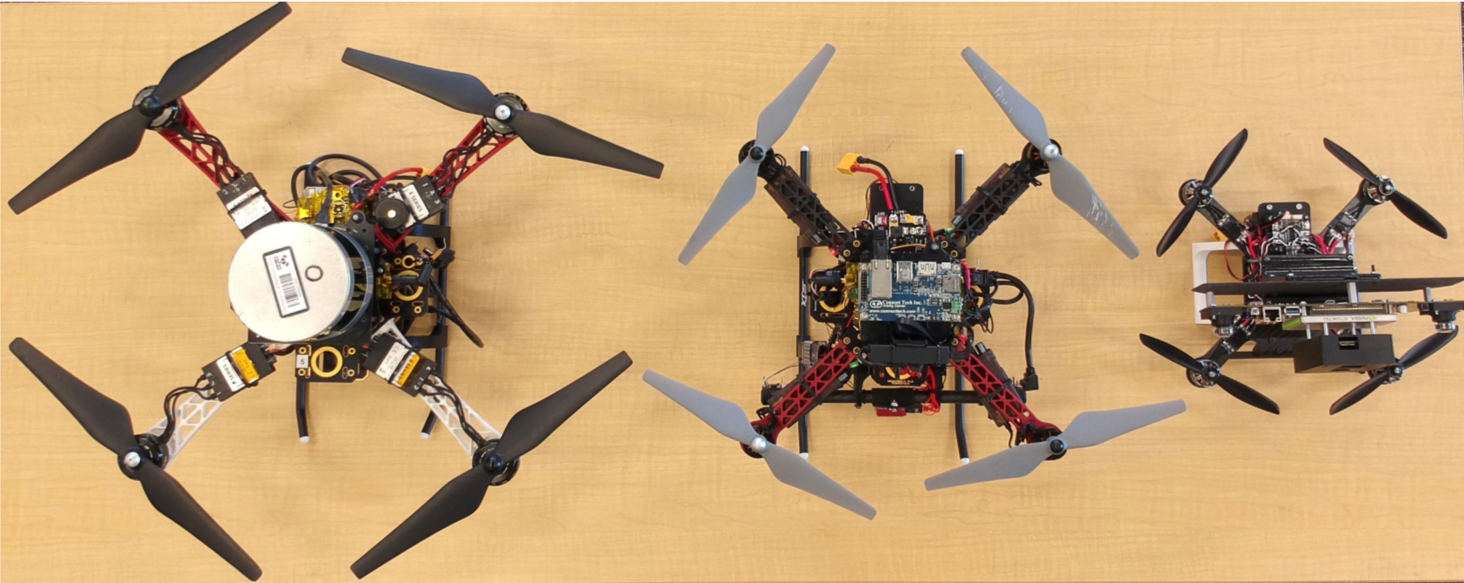}
    \caption{Evolution of the Falcon UAV platform from the Falcon 450 that measured \SI{0.76}{m} from tip to tip to the Falcon 250 that measures \SI{0.40}{\meter} tip to tip. The smaller size necessitated the increasing levels of integration achieved with the Open Vision Computer.}
    \label{fig:Falcon_evolution}
\end{figure}

Throughout our development efforts of the past few years, we have sought to continually shrink the vehicle size to allow flight into more constrained and cluttered environments. Figure \ref{fig:Falcon_evolution} shows several generations of quadrotor systems that we developed in the context of the DARPA Fast Lightweight Autonomy (FLA) program, trending towards smaller and lighter vehicles over time. The leftmost aircraft is referred to as the Falcon 450, so named for the 450 mm span between the motors along each diagonal. This platform was configured with a LIDAR system for obstacle detection, a pair of discrete Point Grey cameras, a Navtech IMU, a laser altimeter and a Pixhawk autopilot unit. All of the sensors were interfaced to an Intel NUC7i7BNH computer system. As shown in~\cite{mohta2018fast} this system could successfully navigate in outdoor environments and through large openings, but its size and weight made it unsuitable for use in typical dwellings where limited door and window apertures required a smaller, more agile aircraft.

Our experience with the Falcon 450 platform informed the design of subsequent smaller aircrafts including the Falcon 250 platform shown on the right of Figure 1. This platform measures \SI{250}{mm} from motor to motor and can easily navigate indoor environments. This agility comes at a cost in terms of payload, where the Falcon 450 could carry on the order of \SI{2}{kg} of payload. The total sensor and compute payload of the Falcon 250 had to be under \SI{1}{kg}.

\begin{figure*}[t]
	\centering
	\includegraphics[width=0.9\linewidth]{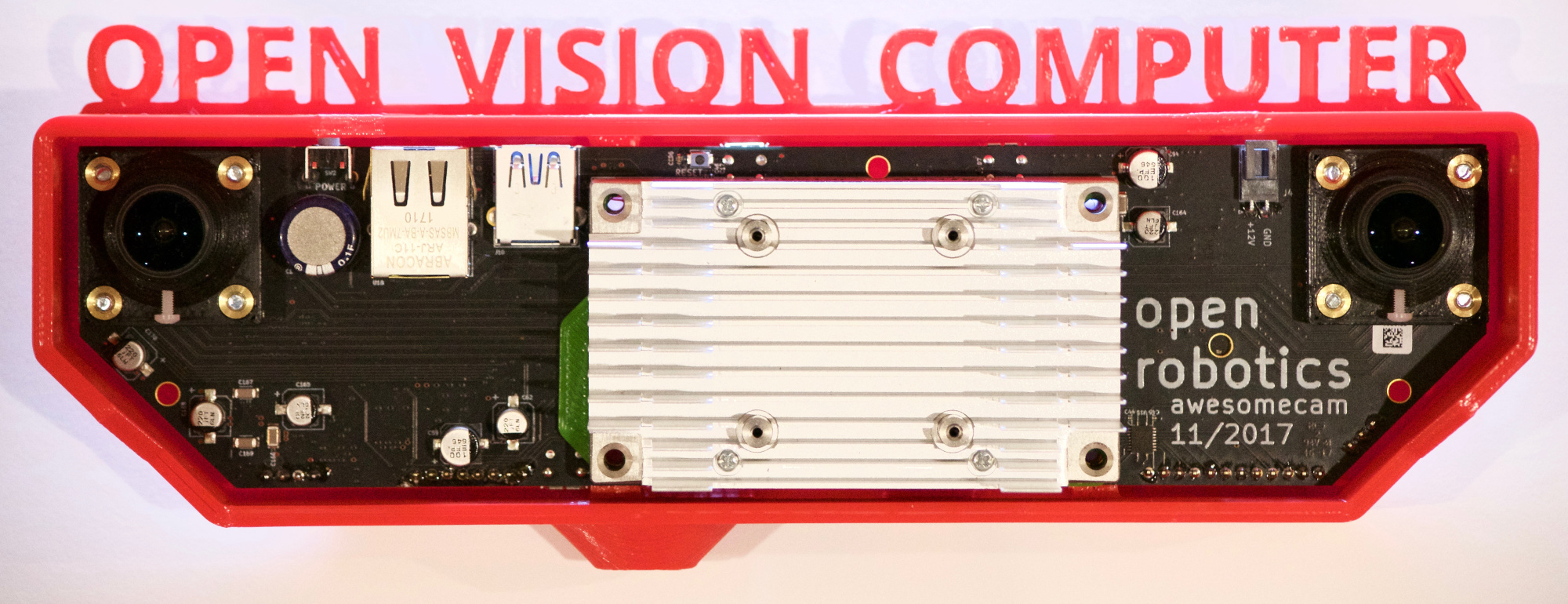}
    \caption{An OVC1 system in a protective plastic shell. The NVIDIA TX2 module is at center. This version of the OVC uses a 20cm baseline between the imagers at far left and right. The FPGA and IMU are on the back side of the PCB.}
    \label{fig:OVC1}
\end{figure*}

On our baseline Falcon 450 system, all sensors and computers were discrete elements connected together via a cabling harness that provided communication and synchronization services. This approach had several drawbacks:
\begin{itemize}
    \item The cabling involved was both complex and a frequent source of failures, particularly in the context of a flying platform that was constantly vibrating.
    \item The added weight induced by the cabling and requisite shielding was quite significant.
    \item The task of synchronizing 2 cameras and an IMU with a coherent auto-exposure algorithm via the Intel NUC added additional complexity to our software system.
\end{itemize}
These considerations led us to develop an integrated system, the Open Vision Computer (OVC), where the major sensing and computing elements are integrated into a single unit to reduce size, weight and complexity of our overall system.

\subsection{Open Source}

As much as possible, the design is intended to be open. Specifically, the schematics, PCB layouts, bill of materials, FPGA firmware, Linux kernel and userland driver source code are available online and released under permissive licenses: \url{http://open.vision.computer}

The electrical design and circuit board layout were accomplished using KiCAD, a cross-platform open source design automation suite~\footnote{\url{http://kicad-pcb.org}}. As such, the design can be viewed and modified without requiring proprietary tool licenses.

Source code of the higher level software, including the Visual Inertial Odometry system, is also released under an open source license, so that others may develop on or learn from our efforts.

\subsection{Related Work}

Several authors have proposed hardware systems for accelerating or embedding computer vision computations for robotic applications. Nikolic et al. \cite{VI-sensor-ICRA14} describe the VI-Sensor which also integrates a pair of image sensors and an inertial measurement unit on a single device along with a field programmable gate array (FPGA) that is used to synchronize all three sensors and to perform low-level image processing operations directly on the image stream. This system is usually employed as a smart sensor that is typically integrated with a high level computer system which performs any higher level steps of the algorithm. Zhang et al. \cite{PIRVS} describe the PIRVS system which  also incorporates a pair of imagers and IMU and a quad core ARM chip into a single system. A single PIRVS unit can perform all aspects of the SLAM pipeline and deliver pose estimates as a service to other applications. This system does an excellent job of integrating the resources required for the SLAM problem into a single low power package, but it does not provide a lot of additional computational capacity for running other applications on board.

The system described in this paper was designed to support a range of vision based applications including visual inertial odometry and stereo, along with all of the other computations required for autonomy such as map construction, path planning, and control. Furthermore, as we will describe in subsequent sections, its considerable user-programmable CPU, GPU, and FPGA resources allow for applications including ``deep learning'' classification tasks, dense stereo reconstruction, mapping, and path planning, all simultaneously running on the same platform.
\section{Architecture of the Open Vision Computer}
\label{sec:architecture}

\begin{figure}[t]
	\centering
	\includegraphics[width=\linewidth]{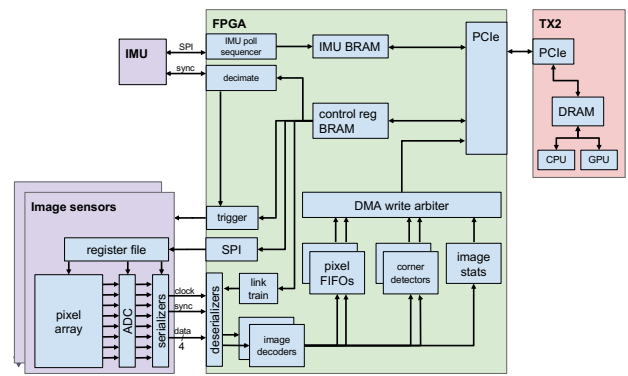}
    \caption{Block Diagram showing the major elements of the Open Vision Computer.}
    \label{fig:OVC_block_diagram}
\end{figure}

Figure \ref{fig:OVC_block_diagram} diagrams the major components of the Open Vision Computer, the first version of which is shown in Figure~\ref{fig:OVC1}. The sensor suite consists of a pair of Python 1300 CMOS image sensors and a VectorNav VN-100 Inertial Measurement Unit (IMU). The image sensors are half-inch format with a resolution of 1280$\times$1024 pixels and feature an electronic global shutter, which is critical for avoiding distortion on fast-moving and fast-rotating aerial vehicles. The experiments described in this paper use the monochrome version of these image sensors, but pin-compatible Bayer color versions are readily available. The VectorNav VN-100 provides the system with temperature-calibrated acceleration and rotational rate measurements along all three axes.

All of the sensors are directly connected to an Intel Cyclone V GT Field Programmable Gate Array (FPGA). The FPGA synchronizes both of the image sensors to the inertial measurement unit and adds hardware timestamps to all measurements, so they can then be processed in software on a common timebase. The FPGA bridges all sensor streams to an NVIDIA TX2 computer module via a 4-lane PCI Express Generation 2.0 bus (PCIe Gen 2.0 x4). This interface provides a theoretical maximum of 16 gigabits per second of full-duplex bandwidth between the FPGA and the TX2.

The NVIDIA TX2 is a single board computer module designed for embedded applications that require high performance computing \cite{franklin2017nvidia}. In addition to a 256-core Pascal GPU, it also offers 4 ARMv8 64-bit A57 CPU cores and 2 ARMv8 64-bit "Denver" cores. The module offers a wide range of peripheral interfaces, including Gigabit Ethernet, USB 3 and USB 2, SPI, multiple UARTs, and a built in 802.11ac WiFi modem.

Using the PCIe bus to interface the sensor subsystem to the compute module offers a number of advantages. First, it offers a directly wired interface that does not consume any of the other interface ports. Second, on previous systems which relied extensively on USB, we noted that the USB cables and hubs that were used to interface various components added nontrivial weight and were a common cause of system failures on systems that were occasionally subjected to violent accelerations or decelerations. Third, extensive USB cabling can also lead to difficult-to-debug EMI leakage that can corrupt GPS receivers and other radios, particularly on platforms subject to repeated high accelerations and crashes which can degrade cable and connector structural integrity over time.

In contrast, the PCIe interface in the OVC is hard-wired and contained on PCB traces and rigid board-to-board connectors. PCIe provides a direct, high speed interface to the NVIDIA TX2 unified memory system that is shared between the CPUs and the GPU. These transfers are made via Direct Memory Access (DMA) and are quite efficient, requiring little CPU overhead.

The TX2 hardware and recent Linux kernels allow re-scanning the PCIe system while the system is running. Because the TX2 has a proprietary system bus for all internal peripherals, the PCIe bus only consists of the TX2 root controller and our FPGA endpoint. As such, it is possible to re-enumerate the PCIe bus by simply unloading and re-loading its kernel module, without requiring the entire operating system to shutdown. This allows us to reconfigure the FPGA system after the operating system is launched. Furthermore, this means that it is not necessary to store the entire FPGA configuration in a special purpose ROM; we can instead store the FPGA configuration in normal files in the operating system and reload as needed at runtime.

In addition to synchronizing the sensors and interfacing them to the TX2 computer module, the FPGA also plays an important role as a computational element. Many researchers have noted that FPGAs with their reconfigurable, highly parallel computational fabric, are well suited to perform real time computation on streaming data sources like video or audio inputs. We leverage this capability by programming the FPGA to perform feature point detection on our input image streams at frame rate. 

Specifically, we have implemented a variant of the Accelerated Segment Test (AST) algorithm \cite{AST} to detect corner features. Corner detection is a crucial first step in the overall visual-inertial odometry (VIO) pipeline. Since this step involves interrogating every pixel in every image, it often represents a significant percentage of the overall computational effort. By implementing this algorithm on the FPGA, we are able to offload this task from the CPU and perform it more efficiently with zero latency, since the features are extracted while the image is streaming from the image sensors on their way to TX2 system memory. A ``feature list'' can then be read by the CPU and GPU immediately following the raw images in RAM.

\begin{figure}
	\centering
	\includegraphics[width=\linewidth]{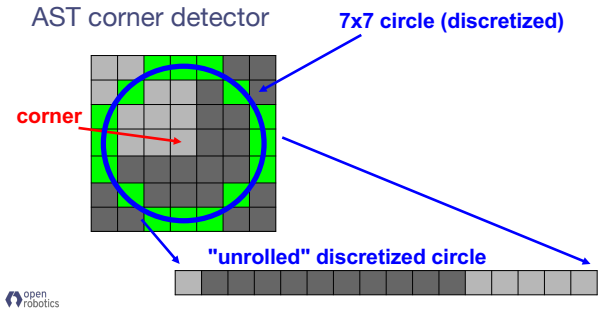}
    \caption{The AST corner detection scheme is implemented on the FPGA to operate on the streaming pixel data by maintaining a rolling buffer of seven lines of the image and constructing a custom computational block to conduct the requisite analysis.}
    \label{fig:AST}
\end{figure}

\begin{figure}
	\centering
	\includegraphics[width=\linewidth]{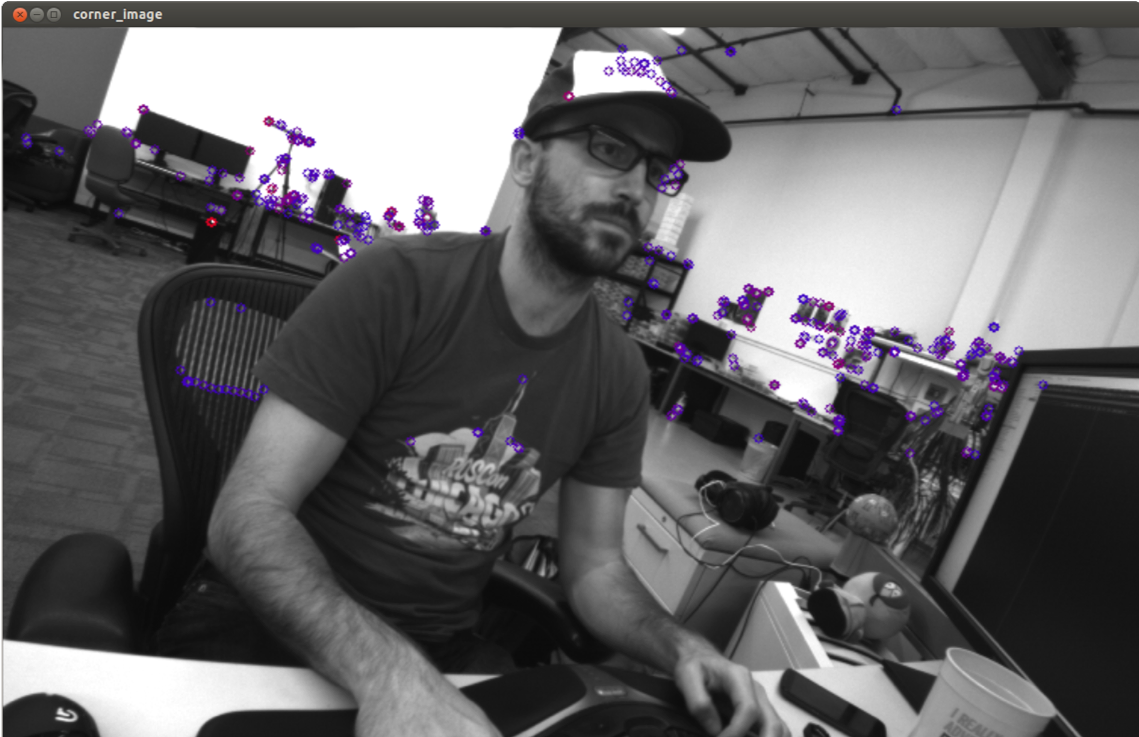}
    \caption{Output of the corner detection procedure on a typical image}
    \label{fig:Morgan_corners}
\end{figure}

Figure \ref{fig:AST} diagrams the core computation behind the AST detector. For each pixel in the image, the system implicitly considers the perimeter of a circle embedded in a 7$\times$7 window surrounding the candidate pixel. First, the intensity of the center pixel is subtracted from the discretized circle. Then, if 9 or more contiguous elements of the 16 entries of the ``difference circle'' are above a threshold value, the pixel is considered a corner candidate. Both ``bright'' and ``dark'' corners are handled simultaneously by maintaining sign bits throughout the computation.

A score is assigned to each corner candidate based on the contrast between the center pixel and the minimum difference to an element of the discretized circle. A non-maximal suppression step is applied to a 3$\times$3 neighborhood to produce distinct, well localized, corner features as in \cite{FAST}. All of these steps are carried out on the FPGA fabric through a rolling buffer of 7 lines of pixels and 3 lines of corner candidate scores. As an implementation detail, the image sensors incorporate four LVDS lanes, resulting in four pixels being decoded on each clock cycle of the deserializer bank. As a result, the corner detector must process four adjacent pixels on each clock cycle. This is accomplished by simply replicating the corner detection unit within the FPGA and picking out four adjacent discretized circles from the rolling window buffer. Similarly, the logic is replicated twice in order to detect corners on both image sensors simultaneously. As a result, eight candidate pixels are evaluated every clock cycle. The computation is heavily pipelined in order to be able to start processing a new set of eight pixels on every clock cycle.

As mentioned previously, the FPGA can be reprogrammed by the system CPU at any time, which means that it can be reconfigured to perform a wide variety of different operations on the incoming pixel stream. Operations including edge detection, image pyramid construction, or frame differencing are just some of the procedures that could be carried out efficiently on an FPGA, although we have not implemented them at time of writing. The last operation, frame differencing, is possible because of the low latency and high bandwidth offered by the PCIe link between the FPGA and the TX2 system memory, allowing for frame buffers and other intermediate computational steps which need significant memory storage.

The first version of the Open Vision Computer (OVC1) is shown in Figure~\ref{fig:OVC1}. This version of the system is configured with a 20 cm baseline between the two imagers. The imagers are equipped with M12 lens mounts, which permits easily interchangeable lenses depending on the needs of the application. The overall system consists of a single 6-layer circuit board \SI{23}{cm} wide and \SI{5.6}{cm} high and weighs less than \SI{200}{grams} with the TX2 module installed. Total power consumption is less than \SI{20}{watts} in practice.
\section{Experiments}

\begin{figure}
	\centering
	\includegraphics[width=0.8\linewidth]{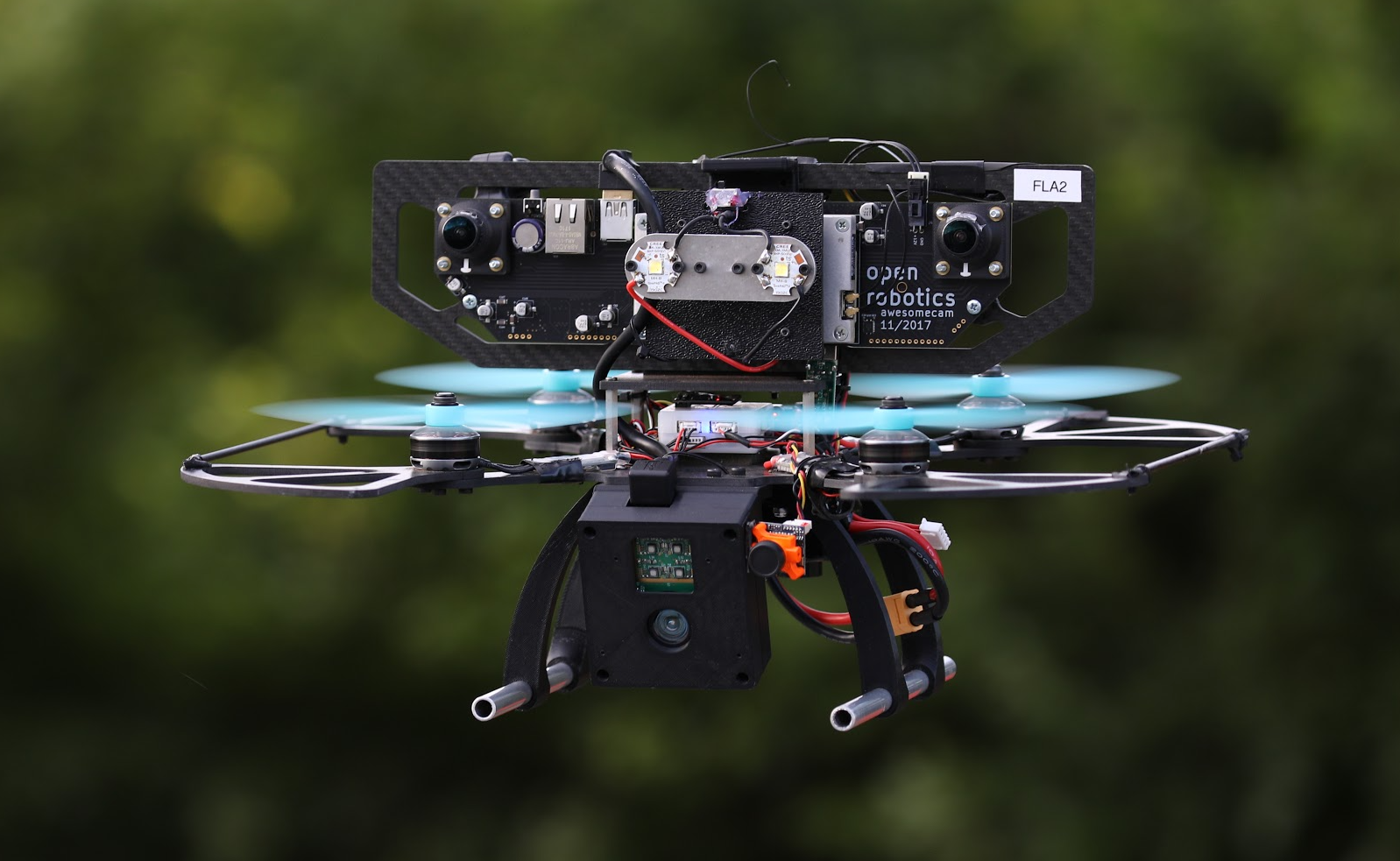}
    \caption{The Falcon 250 autonomous flying robot. This system is equipped with an Open Vision Computer, a PMD Monstar depth camera, a laser altimeter and a Pixhawk flight controller.}
    \label{fig:Falcon250}
\end{figure}

The overall goal of the project was to develop an integrated sensing and computational unit that could be deployed on our Falcon 250 autonomous flying robot shown in Figure \ref{fig:Falcon250}.

This airframe has an all up weight of \SI{1.3}{kg} and is equipped with an Open Vision Computer, which serves as the computational engine for the system, a PMD Monstar range camera, a laser altimeter system and a Pixhawk flight controller, which performs low-level flight control functions. The PMD Monstar range sensor provides a 352$\times$287 range image over a 100 degree horizontal field of view out to a range of approximately \SI{7}{meters} in indoor environments.

The system is designed to perform fully autonomous navigation and exploration operations in GPS-denied indoor and outdoor environments. One class of missions that was performed with this platform involved giving the system a direction, a heading and a maximum altitude and asking it to navigate to that location and return. The system was able to successfully traverse hundreds of meters avoiding trees, buildings and other obstacles and return to the start position without using any global position information such as GPS or other radio beacons.

\begin{figure*}[t]
\centering
\begin{tabular}{cccc}
\includegraphics[width=0.22\linewidth]{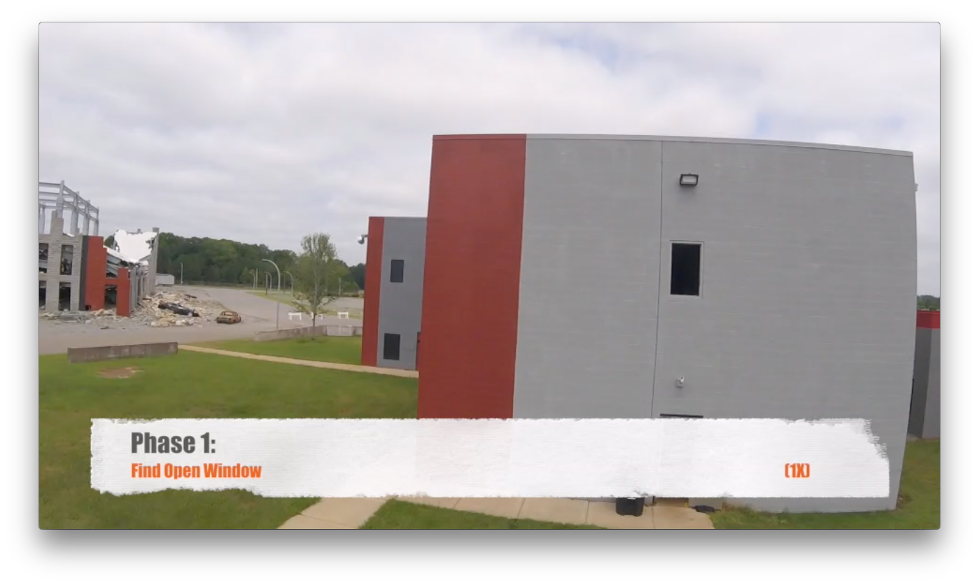} &
\includegraphics[width=0.22\linewidth]{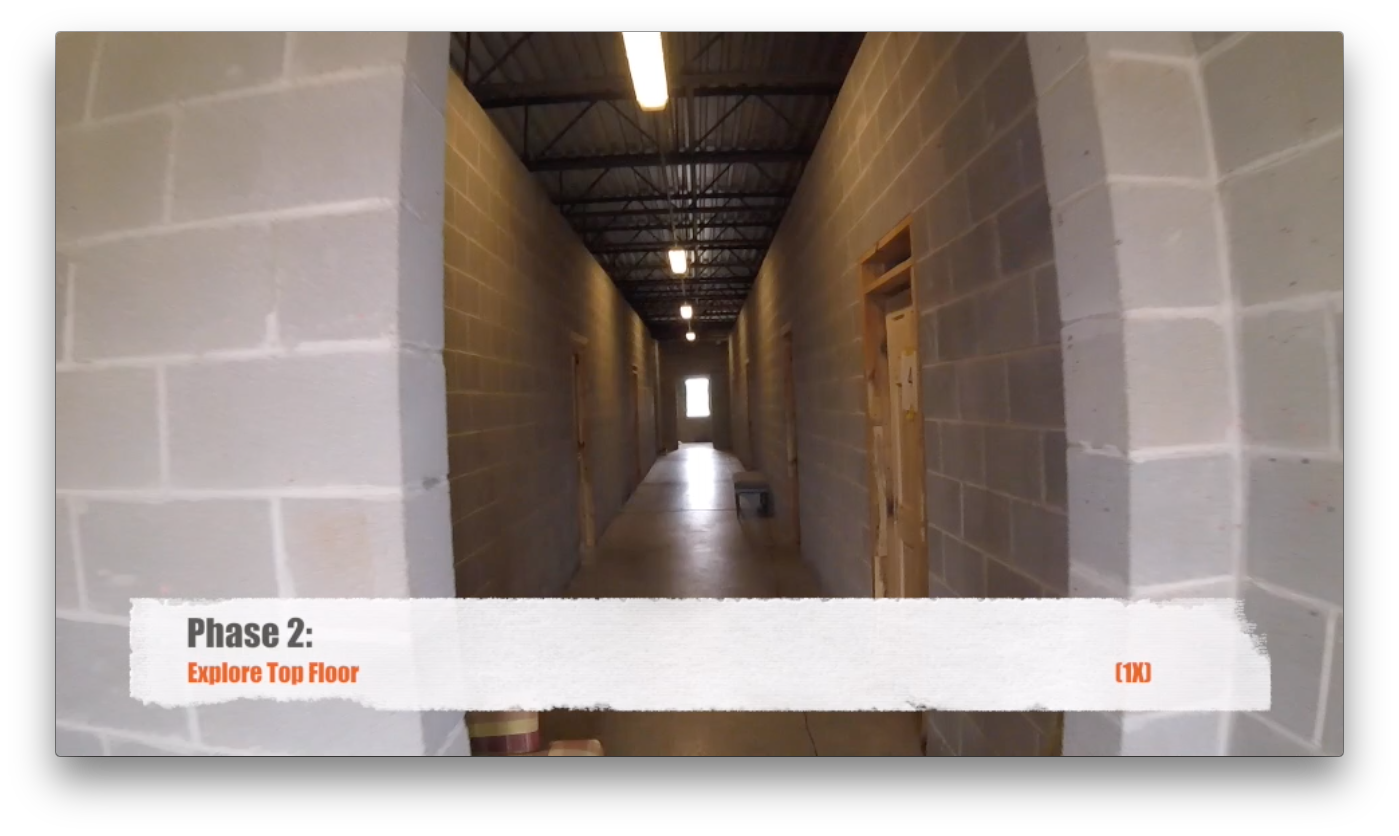} &
\includegraphics[width=0.22\linewidth]{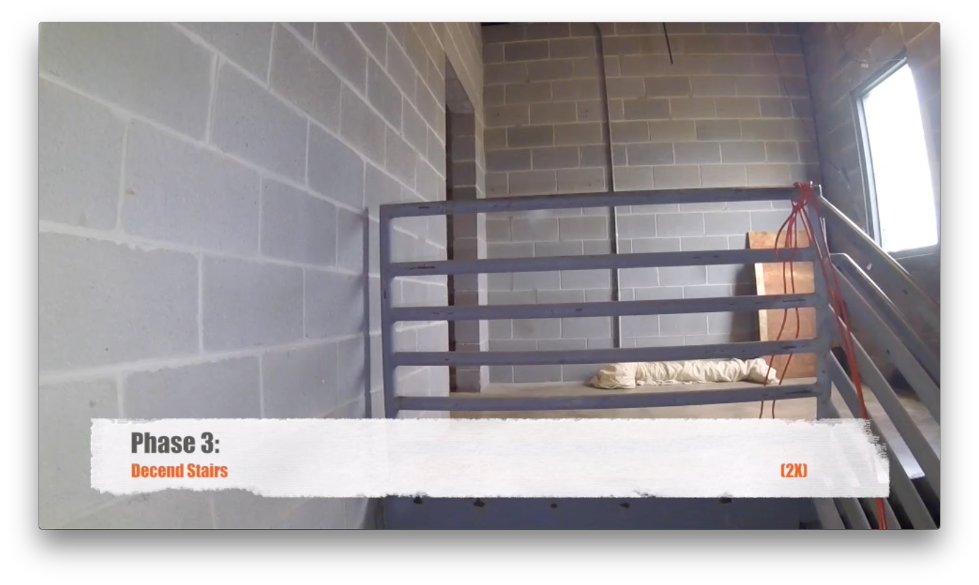} &
\includegraphics[width=0.22\linewidth]{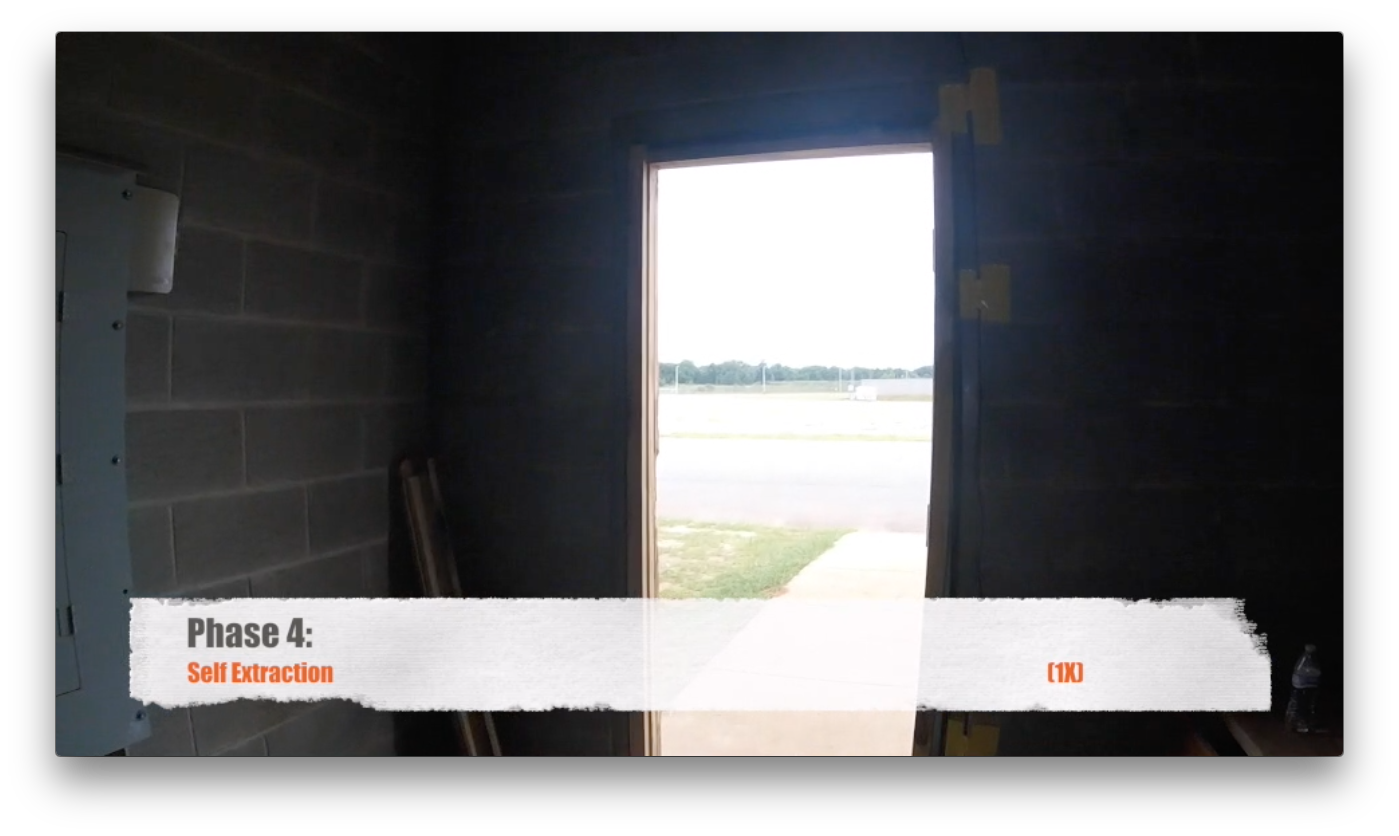} \\
1 & 2 & 3 & 4
\end{tabular}
\caption{Example of an autonomous mission composed of four phases, 1) finding and entering an open second floor window 2) exploring the second floor 3) finding and descending a staircase 4) finding and exiting a first floor doorway. }
\label{fig:phases}
\end{figure*}

The other class of tasks that were performed involved having the robot autonomously explore indoor environments. More specifically, the robot was initially placed outside and downrange of a target building. The high level mission plan involved a series of vaguely-defined steps: approach the building, find an open second floor window, enter the window, explore the second floor, find a staircase, descend the staircase, and exit the building. All phases of the mission were carried out by the robot using only the computation and sensing available on the platform. Figure \ref{fig:phases} shows stills from various phases of this mission.

In order to perform these kinds of tasks, the robot needed to be able run a range of processes in parallel in real time. First, a real-time stereo visual inertial odometry (VIO) process was implemented so that the robot could maintain an estimate of its position. A real-time GPU accelerated stereopsis procedure was used to provide an estimate of obstacle locations relative to the robot, which was particularly relevant when the system was operating in outdoor environments where the PMD depth camera was inoperative due to interference from the sun \cite{hernandez2016embedded}. Depth measurements from the stereopsis and PMD sensors could be fused with the position estimate to form a three dimensional map of the environment as the robot moved through the scene \cite{kuhnert2006fusion,gudmundsson2008fusion,hahne2008combining,zhu2008fusion}. An advanced motion primitive based planner was continually executed to plan and replan the robots trajectory in three dimensions as its understanding of its surroundings changed. Finally, a high level state machine controller was used to execute and monitor the overall mission. More detailed descriptions of the specific computational modules including the visual odometry system and real time planner can be found in the following publications \cite{mohta2018fast,svacha2017improving,liu2017search,sun2018robust,watterson2018control}.
All of the various components of the application were integrated and coordinated using the Robot Operating System (ROS) \cite{ros}. 

Table \ref{tab:processes} lists the major processes that were running on the robot and indicates which computational unit or units on the OVC performed the work along with the rate in Hz at which the process was running.

\begin{table}
\caption{Processes running on OVC during a typical exploration mission.}
\label{tab:processes}
\centering
\begin{tabular}{l c c}
\toprule
\textbf{Process} & \textbf{Computation required} & \textbf{Rate (Hz)} \\
\midrule
Stereo VIO & FPGA + 1 ARM Cores & 20 \\

UKF State Estimation & 0.3 ARM Cores & 200 \\ 

3D Map Making & 0.5 ARM Cores & 10 \\

3D Planner & 0.5 ARM Cores & 3 \\ 

Controller and State Machine & 0.5 ARM Cores & 200 \\ 

Stereo (SGM) & GPU + 0.2 ARM Cores & 2 \\ 
\bottomrule
\end{tabular}
\end{table}

Note that the computational stack described in this section utilizes all aspects of the OVC device: the CPU cores, the GPU, and the FPGA. We note that this workload is typical of many robotic applications which are inherently parallel with multiple concurrent real-time processes running and interacting. These kinds of applications benefit from the architecture of the OVC which provides multiple heterogeneous computational capabilities, each of which is best suited to particular types of data processing. The strengths and weaknesses of using the CPU, GPU, and FPGA for each subtask can be balanced as appropriate in a multi-dimensional tradeoff between required throughput, latency and jitter tolerance, intrinsic computational parallelism (or lack thereof), and ease of programming.
\section{Current and Future Work}

This section describes several areas of ongoing and future work planned using the Open Vision Computer.

\begin{figure}
	\centering
	\includegraphics[width=\columnwidth]{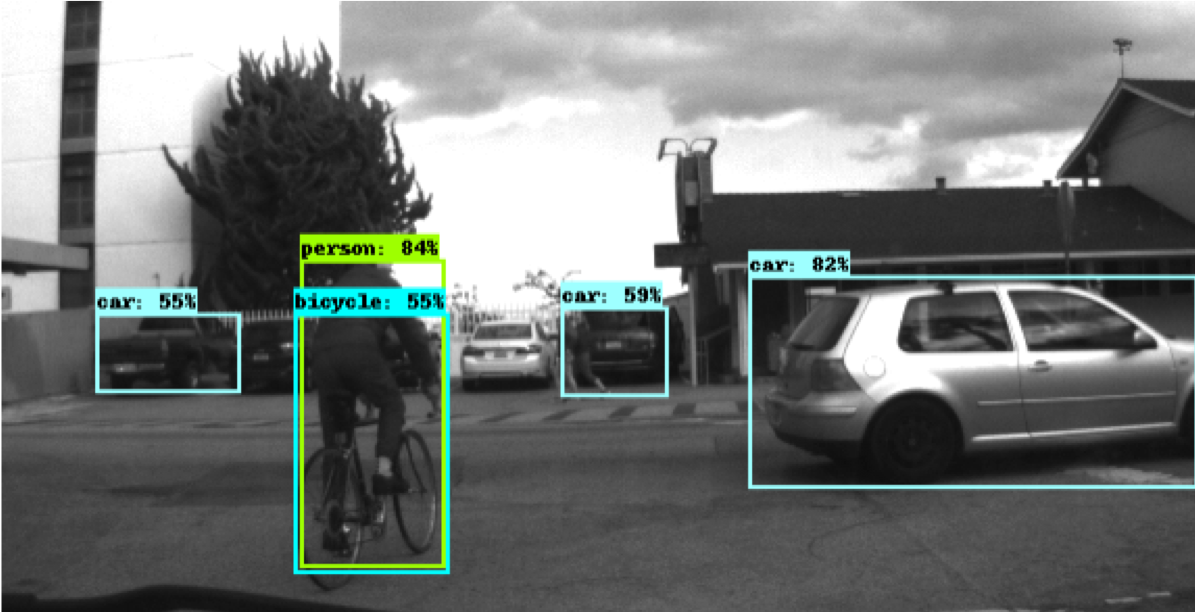}
    \caption{Typical output of the SSD 512 network using TensorFlow, running at 10 Hz on the system's GPU and CPU.}
    \label{fig:ssd}
    \vspace{-0.3cm}
\end{figure}

\subsection{Object Recognition and Segmentation}

As mentioned in Section~\ref{sec:architecture}, the NVIDIA TX2 module contains a sizable GPU in addition to the CPU core cluster. This makes the module well-suited for the current generation of deep learning applications, including object detection and semantic segmentation.  To validate the capability of the system for object recognition using  ``deep learning`` techniques, we ran TensorFlow using the Single Shot MultiBox Detector (SSD512), balancing the load across the GPU and CPU resources and processing images arriving via DMA from the FPGA. The resulting network runs at approximately 10 frame/sec while consuming \SI{12}{watts} of electrical power. Typical output is shown in Figure~\ref{fig:ssd}.

We have also successively run optimized segmentation networks such as ERFNet for semantic segmentation tasks \cite{romera2018erfnet}, achieving frame rates of approximately 6 frame/sec with the original network using images of resolution 640$\times$512, and upto 10 frame/sec on our modified version of this network for semantic segmentation of 2-3 classes.

\subsection{OVC Rev 2}

\begin{figure}
	\centering
	\includegraphics[width=0.8\columnwidth]{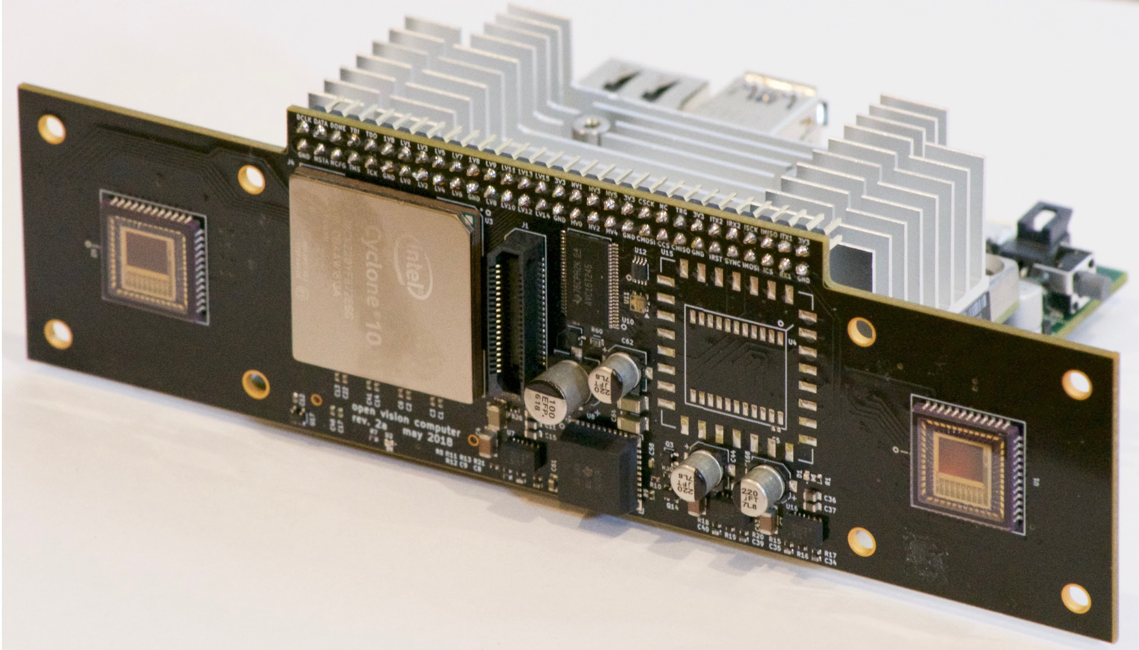} \\
	\vspace{0.5 em}
	\includegraphics[width=0.8\columnwidth]{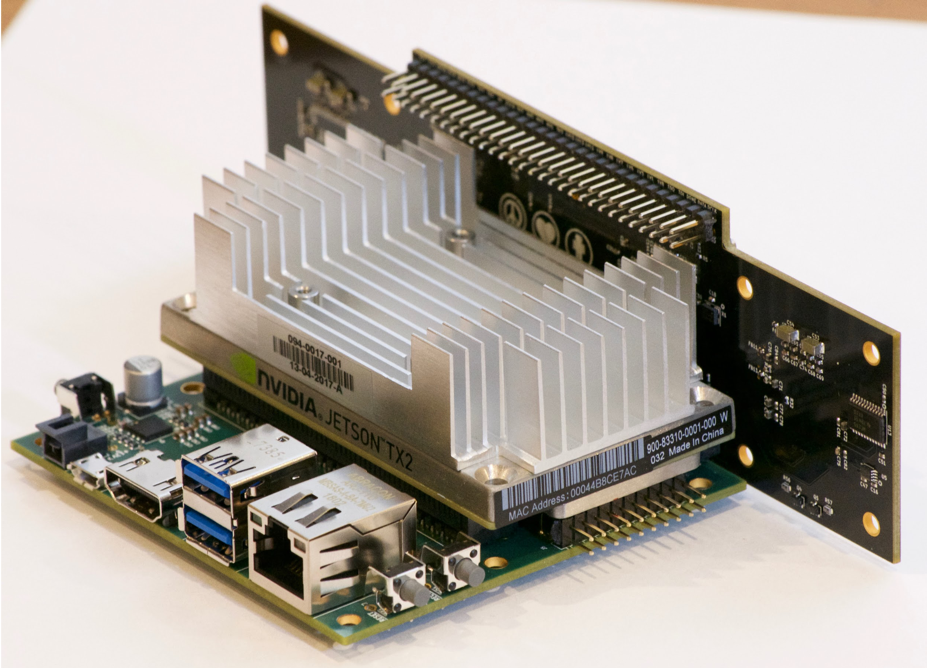}
    \caption{A follow-on design, the OVC2, aims to shrink the form factor and improve performance.}
    \label{fig:ovc2}
    \vspace{-0.5cm}
\end{figure}

We are incorporating many lessons learned from the design of OVC1 into a successor project, following the same design concept of building around an NVIDIA TX2 and a FPGA over PCIe x4. The most significant change is to upgrade the FPGA to an Intel Cyclone 10 GX. This FPGA includes many architectural improvements, including hard floating-point support in its DSP blocks and improved performance across the board thanks to fabrication process improvements.

To permit usage in more space-constrained applications, we have reduced the image sensor baseline to \SI{12}{cm} and split the circuitry into two boards: one board contains the image sensors, inertial measurement unit, and FPGA, and the other board contains the TX2 module and its associated connectors and supporting circuitry. These two boards are connected via a high speed 90 degree connector which carries the PCIe signals. Partitioning the design at the PCIe bus provides a well defined, standards based interface between the two subsystems.

There are several advantages to this design. First, it provides us with a smaller, more aerodynamic profile, as shown in Figure~\ref{fig:ovc2} as compared to Figure~\ref{fig:OVC1} - the entire OVC2 system is \SI{16}{cm} wide, \SI{4.5}{cm} tall and \SI{11}{cm} deep and weighs under \SI{200}{grams}. Second, it cleanly separates the sensor components from the computational components, which allows for further designs where different choices are made on either side. For example, one could design a new sensor board using a single monocular camera, a different Inertial Measurement Unit, or a completely different set of sensors without having to redo the layout of the compute board. Similarly, as technologies evolve we plan to take advantage of new single board computer systems such as the recently announced NVIDIA Xavier system which offers much higher performance than the TX2. Here, we can leverage much of the experience gained and software developed for the TX2 platform to more quickly develop increasingly functional systems derived from, or even directly using, the PCIe-based FPGA and sensor board described in this section.

Additional improvements were added to the OVC2 design such as exposing the TX2 serial ports on the board to allow for direct interfacing to the Pixhawk flight controller, and increased access to General Purpose Input Output (GPIO) pins which can be programmatically controlled by both the FPGA and the TX2 CPU cluster. These pins can be used to provide a variety of signals such as a synchronization pulse linked to the exposure period of the camera which can be used to drive external LEDs that only illuminate the scene during an image exposure. Alternatively they could be used as inputs to synchronize or timestamp external sensors relative to the time base of the cameras and the IMU.

As with OVC1, all aspects of the OVC2 design are freely available on the Internet and can be downloaded from \url{http://open.vision.computer}
\section{Conclusion}

In this paper, we presented the design of the Open Vision Computer, a self-contained platform featuring leading-edge embedded CPU, GPU, and FPGA resources with tightly coupled image and inertial sensors. 
The design of this system was motivated by the goal of developing flying robots that could autonomously navigate through GPS denied indoor and outdoor environments. By tightly integrating the sensor and computational elements into a single package we were able to meet the stringent size, weight and power requirements imposed by the task. Furthermore this integration effort improved the overall reliability of the system since many issues related to cabling failures and sensor mounting were obviated by the all in one design where the sensors were rigidly mounted to each other on a single circuit board.

The OVC system offers multiple computational resources that can operate in parallel including a reconfigurable Field Programmable Gate Array, a 256 core GPU and 6 ARM cores. This arrangement allows us to tackle different elements of our problem using computational and implementation strategies best suited to those workloads. Consequently our robot is able to perform multiple concurrent processes in a highly efficient manner while using less than 20 Watts of power.
We have demonstrated the utility of the Open Vision Computer by using it as the primary computational system on a range of successful autonomous flights through challenging environments on our Falcon 250 UAV.

We have made the design of the Open Vision Computer and the associated software available as an open source project to allow others to build on or learn from our work.

While the OVC was developed with the needs of flying robots uppermost, the system can be readily employed on a variety of other robotic platforms including wheeled and legged vehicles which we are now demonstrating in other projects. Here the fact that the system can be integrated simply by supplying a power cable and an Ethernet or WiFi connection allows us to easily port much of the software and capabilities we have developed for flying systems onto other robotic systems.

\bibliographystyle{IEEEtran}
\bibliography{OVC}

\begin{thebibliography}{10}
\providecommand{\url}[1]{#1}
\csname url@rmstyle\endcsname
\providecommand{\newblock}{\relax}
\providecommand{\bibinfo}[2]{#2}
\providecommand\BIBentrySTDinterwordspacing{\spaceskip=0pt\relax}
\providecommand\BIBentryALTinterwordstretchfactor{4}
\providecommand\BIBentryALTinterwordspacing{\spaceskip=\fontdimen2\font plus
\BIBentryALTinterwordstretchfactor\fontdimen3\font minus
  \fontdimen4\font\relax}
\providecommand\BIBforeignlanguage[2]{{%
\expandafter\ifx\csname l@#1\endcsname\relax
\typeout{** WARNING: IEEEtran.bst: No hyphenation pattern has been}%
\typeout{** loaded for the language `#1'. Using the pattern for}%
\typeout{** the default language instead.}%
\else
\language=\csname l@#1\endcsname
\fi
#2}}

\bibitem{mohta2018fast}
K.~Mohta, M.~Watterson, Y.~Mulgaonkar, S.~Liu, C.~Qu, A.~Makineni, K.~Saulnier,
  K.~Sun, A.~Zhu, J.~Delmerico, \emph{et~al.}, ``Fast, autonomous flight in
  gps-denied and cluttered environments,'' \emph{Journal of Field Robotics},
  vol.~35, no.~1, pp. 101--120, 2018.

\bibitem{VI-sensor-ICRA14}
J.~Nikolic, J.~Rehder, M.~Burri, P.~Gohl, S.~Leutenegger, P.~T. Furgale, and
  R.~Siegwart, ``A synchronized visual-inertial sensor system with fpga
  pre-processing for accurate real-time slam,'' in \emph{2014 IEEE
  International Conference on Robotics and Automation (ICRA)}, May 2014, pp.
  431--437.

\bibitem{PIRVS}
Z.~Zhang, S.~Liu, G.~Tsai, H.~Hu, C.-C. Chu, and F.~Zheng, ``Pirvs: An advanced
  visual-inertial slam system with flexible sensor fusion and hardware
  co-design,'' in \emph{Proceedings of the International Conference on Robotics
  and Automation}, 2018.

\bibitem{franklin2017nvidia}
D.~Franklin, ``Nvidia jetson tx2 delivers twice the intelligence to the edge,''
  \emph{NVIDIA Accelerated Computing| Parallel Forall}, 2017.

\bibitem{AST}
E.~Mair, G.~D. Hager, D.~Burschka, M.~Suppa, and G.~Hirzinger, ``Adaptive and
  generic corner detection based on the accelerated segment test,'' in
  \emph{Computer Vision -- ECCV 2010}, K.~Daniilidis, P.~Maragos, and
  N.~Paragios, Eds.\hskip 1em plus 0.5em minus 0.4em\relax Berlin, Heidelberg:
  Springer Berlin Heidelberg, 2010, pp. 183--196.

\bibitem{FAST}
E.~Rosten and T.~Drummond, ``Machine learning for high-speed corner
  detection,'' in \emph{European Conference on Computer Vision}, 2006.

\bibitem{hernandez2016embedded}
D.~Hernandez-Juarez, A.~Chac{\'o}n, A.~Espinosa, D.~V{\'a}zquez, J.~C. Moure,
  and A.~M. L{\'o}pez, ``Embedded real-time stereo estimation via semi-global
  matching on the gpu,'' \emph{Procedia Computer Science}, vol.~80, pp.
  143--153, 2016.

\bibitem{kuhnert2006fusion}
K.-D. Kuhnert and M.~Stommel, ``Fusion of stereo-camera and pmd-camera data for
  real-time suited precise 3d environment reconstruction.'' in \emph{IROS},
  2006, pp. 4780--4785.

\bibitem{gudmundsson2008fusion}
S.~A. Gudmundsson, H.~Aanaes, and R.~Larsen, ``Fusion of stereo vision and
  time-of-flight imaging for improved 3d estimation,'' \emph{International
  Journal of Intelligent Systems Technologies and Applications}, vol.~5, no.
  3-4, pp. 425--433, 2008.

\bibitem{hahne2008combining}
U.~Hahne and M.~Alexa, ``Combining time-of-flight depth and stereo images
  without accurate extrinsic calibration,'' \emph{International Journal of
  Intelligent Systems Technologies and Applications}, vol.~5, no. 3-4, pp.
  325--333, 2008.

\bibitem{zhu2008fusion}
J.~Zhu, L.~Wang, R.~Yang, and J.~Davis, ``Fusion of time-of-flight depth and
  stereo for high accuracy depth maps,'' in \emph{Computer Vision and Pattern
  Recognition, 2008. CVPR 2008. IEEE Conference on}.\hskip 1em plus 0.5em minus
  0.4em\relax IEEE, 2008, pp. 1--8.

\bibitem{svacha2017improving}
J.~Svacha, K.~Mohta, and V.~Kumar, ``Improving quadrotor trajectory tracking by
  compensating for aerodynamic effects,'' in \emph{Unmanned Aircraft Systems
  (ICUAS), 2017 International Conference on}.\hskip 1em plus 0.5em minus
  0.4em\relax IEEE, 2017, pp. 860--866.

\bibitem{liu2017search}
S.~Liu, N.~Atanasov, K.~Mohta, and V.~Kumar, ``Search-based motion planning for
  quadrotors using linear quadratic minimum time control,'' in
  \emph{Intelligent Robots and Systems (IROS), 2017 IEEE/RSJ International
  Conference on}.\hskip 1em plus 0.5em minus 0.4em\relax IEEE, 2017, pp.
  2872--2879.

\bibitem{sun2018robust}
K.~Sun, K.~Mohta, B.~Pfrommer, M.~Watterson, S.~Liu, Y.~Mulgaonkar, C.~J.
  Taylor, and V.~Kumar, ``Robust stereo visual inertial odometry for fast
  autonomous flight,'' \emph{IEEE Robotics and Automation Letters}, vol.~3,
  no.~2, pp. 965--972, 2018.

\bibitem{watterson2018control}
M.~Watterson and V.~Kumar, ``Control of quadrotors using the hopf fibration on
  so (3),'' in \emph{Proceedings of the 2017 International Symposium on
  Robotics Research}, 2018.

\bibitem{ros}
M.~Quigley, B.~Gerkey, K.~Conley, J.~Faust, T.~Foote, J.~Leibs, E.~Berger,
  R.~Wheeler, and A.~Ng, ``Ros: an open-source robot operating system,'' in
  \emph{Proceedings of the Open-Source Software Workshop of the International
  Conference on Robotics and Automation}, 2009.

\bibitem{romera2018erfnet}
E.~Romera, J.~M. Alvarez, L.~M. Bergasa, and R.~Arroyo, ``Erfnet: Efficient
  residual factorized convnet for real-time semantic segmentation,'' \emph{IEEE
  Transactions on Intelligent Transportation Systems}, vol.~19, no.~1, pp.
  263--272, 2018.

\end{thebibliography}

\balance

\end{document}